\documentclass[a4paper,12pt]{article}
\usepackage[margin=2.5cm,nohead]{geometry}

\usepackage{subcaption} 
\usepackage{graphicx}
\graphicspath{ {figs/} }
\usepackage{float}
\usepackage{indentfirst}

\usepackage[hidelinks, colorlinks=false, pdfpagelabels=false, unicode]{hyperref}

\usepackage{natbib}
\bibpunct[, ]{(}{)}{,}{a}{}{,} 

\usepackage{titlesec}
\titleformat{\paragraph} 
{\normalfont\normalsize\bfseries}{\theparagraph}{1em}{}
\titlespacing*{\paragraph}
{0pt}{3.25ex plus 1ex minus .2ex}{1.5ex plus .2ex}

\usepackage{booktabs}
\usepackage{tabularx}
\usepackage{multicol}
\usepackage{multirow}
\usepackage{verbatim}
\usepackage{setspace}

\title{Actuation without production bias}
\author{James Kirby\textsuperscript{a} and Morgan Sonderegger\textsuperscript{b}\\
\small{\textsuperscript{a} Institute for Phonetics and Speech Processing, Ludwig-Maximilians-Universit\"at M\"unchen}\\
\small{\textsuperscript{b}Department of Linguistics, McGill University}\\ \small{\textsuperscript{a}\textit{jkirby@phonetik.uni-muenchen.de, \textsuperscript{b}morgan.sonderegger@mcgill.ca}}}
\date{Preprint: June 2024}

\begin{document}

\maketitle

\doublespacing

\begin{abstract}
Phonetic production bias is the external force most commonly invoked in computational models of sound change, despite the fact that it is not responsible for all, or even most, sound changes. Furthermore, the existence of production bias alone cannot account for how changes do or do not propagate throughout a speech community. While many other factors have been invoked by (socio)phoneticians, including but not limited to contact (between subpopulations) and differences in social evaluation (of variants, groups, or individuals), these are not typically modeled in computational simulations of sound change. In this paper, we consider whether production biases have a unique dynamics in terms of how they impact the population-level spread of change in a setting where agents learn from multiple teachers. We show that, while the dynamics conditioned by production bias are not unique, it is not the case that all perturbing forces have the same dynamics: in particular, if social weight is a function of individual teachers and the correlation between a teacher's social weight and the extent to which they realize a production bias is weak, change is unlikely to propagate. Nevertheless, it remains the case that changes initiated from different sources may display a similar dynamics. A more nuanced understanding of how population structure interacts with individual biases can thus provide a (partial) solution to the `non-phonologization problem'.

\end{abstract}



\section{Introduction\label{sec:intro}}
How and why do speech sounds change? An outstanding goal of much work on sound change has been to understand what kinds of sound change are and aren't possible, and why a given change takes place in certain languages when the necessary pre-conditions are present, but not in others -- the celebrated \textsc{actuation problem} of \citet{weinreich1968empirical}. If we think of a sound change as characterizing a change in the norm of a speech community (a definition that would certainly not be shared by all researchers), the actuation problem is separable into at least three distinct sub-problems:


    \begin{enumerate}
        \item What conditions the generation of an innovative variant in the speech of a single speaker? 
        \item Why do certain individuals adopt the innovative variants of others?
        \item How do variants propagate in a speech community?
    \end{enumerate}

\noindent We refer to these as the problems of \textsc{initiation}, \textsc{transmission} and \textsc{propagation}, respectively (cf. \citealp{bermudez-otero_initiation_2020, lindblom1995sound,milroy_linguistic_1985,janda2003phonologization,stevens2014individual,croft_explaining_2000,hall-lew_individuals_2021}). 

Implicit in this type of bi- or tripartite\footnote{We find it useful to explicitly separate problems 2 and 3 in order to emphasize the distinction between the transmission of a variant between two individuals (sub-problem 2) and the propagation of a variant throughout a speech community to the extent that it becomes the dominant variant in that community (sub-problem 3). In the limiting case where the speech community in question consists of just two individuals, then these will functionally be the same, but we think it is still useful to draw a distinction between transmission/learning at the individual level and propagation/spread at the population level.} division is the idea that change at the \textit{population} level originates in phonetic variation at the \textit{individual} level, an assumption made explicit by researchers such as \citet[184]{ohala1981listener}: ``...the initiation of such sound changes is accomplished by the phonetic mechanism just described; their spread, however, is done by social means, e.g., borrowing, imitation, etc.''

Again, while not all researchers agree that changes at the individual level are relevant, observable, or even theoretically coherent (see e.g., discussions in \citealp{weinreich1968empirical,hall-lew_individuals_2021}, and \citealp{garrett2015sound}), a great deal of work has implicitly or explicitly assumed some division of labor along these lines. It is thus worth exploring whether a model that assumes these components is in principle capable of generating patterns of change that correspond to the observed typology.

\subsection{Initiation of change: phonetic biases}

A common assumption made in many models is that while on some level sound change involves language learners/users selecting from what Ohala memorably called the `pool of synchronic variation' \citep{ohala1989sound}, the variants in the pool are determined by universal aspects of speech production and perception, or phonetic \textsc{bias factors} (\citealp{ohala1993phonetics,blevins2004evolutionary,blevins2006theoretical,garrett2013phonetic}, and many others). Phonetic bias factors are asymmetries in patterns of phonetic realization which make certain sound changes more likely than others.
Phonetic biases were traditionally understood to be \textsc{production biases} -- physiological constraints on speech production -- but acoustic-perceptual biases may also play a role (for reviews see \citealp{garrett2013phonetic,ohala1997aerodynamics,blevins_evolutionary_2015}).
In this chapter, we use the term \textsc{production bias} to refer specifically to incremental and asymmetric phonetic biases with a articulatory basis. For example, vowel undershoot is the result of a production bias affecting the speed with which the articulators involved can achieve a particular spatial target. This is to be distinguished from vowel reduction, the outcome of phonologizing undershoot. What is important for present purposes is that production bias refers to the constraints giving rise to asymmetric, but non-phonologized variability.

Importantly, the mere existence of phonetic biases does not entail that they will lead to change at the population level. For as long as the existence of external forces impinging on speech patterns has been noted, so too has the observation that they don't make change inevitable: indeed, the equilibrium state of language is arguably stability, not change \citep{milroy1992linguistic,janda2003phonologization,kiparsky2015phonologization}. At a minimum, a bias needs to result in a change in the speech behavior of at least a single individual, the classic Ohalaian `mini-sound change'. The idea that sound changes can take hold in the speech of individuals has led to work investigating why certain individuals (the \textsc{innovators} of \citealp{milroy_linguistic_1985}) may be more poised to adopt bias variants than others. Proposals include individual differences in perception and/or cognitive processing style \citep{yu2013phonetic,yu2019individual}, production \citep{dediu_pushes_2019}, and/or learning experience (\citealp{bermudez-otero_initiation_2020}). For more discussion of the role of individual differences in sound change, see the papers in \citealp{hall-lew_individuals_2021}). The critical point for the present discussion is simply that while biases exist, they do not necessarily \textit{entail} change, even at the level of the individual.

\subsection{Propagation of change: non-phonetic bias}

Much as not all biases result in innovations in the speech of individual speakers, not all innovations propagate in a population. Why not? Here a wide range of proposals have been advanced. \citet{lindblom1995sound} suggest that both articulatory ease and perceptual confusability may contribute to making some changes more or less likely to take hold. 
Other researchers propose that bias factors may be kept in check by a countervailing structural force, such as categoricity bias \citep{kirby2013model,kirby2015bias} or contrast maintenance \citep{soskuthy2015understanding}. \citet{kirby2015bias}, building on the tradition of modeling population-level language change of Niyogi et al.~\citep{niyogi1996language,niyogi2006computational,sonderegger2010combining,sonderegger2013variation}, show how \emph{population dynamics} can be a source of stability: a change present in some individuals rooted in the same bias factor can spread or not spread through the population, depending on the learning algorithm assumed at the level of individuals and on population structure (e.g., how connected agents are), with population-level dynamics differing crucially depending on whether each agent learns from one or multiple teachers \citep{niyogi2006computational,niyogi2009proper,smith2009iterated,kirby2013model,kirby2015bias}.

There is, however, something of a disconnect between computational models of sound change propagation and theoretical and empirical work on change in the population. As noted above, much work on sound change from a phonetic perspective has emphasized the role of phonetic bias factors, and these have been implemented in models such as \citet{pierrehumbert2001exemplar} or \citet{soskuthy2015understanding}. Yet, the broader literature on language variation and change often refers to other forces driving spread of a variant through a population: contact between groups with and without the variant, and social meaning attached to the variant or those who use it. As the above quote from Ohala suggests, the forces which give rise to the initial conditions for change at the level of the individual may be distinct from those which drive transmission and propagation at the population level. To understand the extent to which phonetic biases actually shape sound change typology, then, it is necessary to study them within the context of a population as well as within individuals, and to consider the full spectrum of forces which cause or inhibit the spread of a variant.

Perhaps most prosaically,  
frequency of interaction between and/or accommodation to individuals or \textit{groups} of individuals impacts both transmission and propagation of variants \citep{trudgill_dialects_1986}. Indeed, the many decades of research in the Labovian tradition argues that propagation of change in a speech community involves orderly shifts in the frequency of competing variants along demographic dimensions \citep{labov2001principles}. All else being equal, speakers are more likely to imitate the speech of groups (however defined, e.g., speakers of a different dialect, a certain gender, or a locally-defined category such as `burnout' in \citealp{eckert_language_2000}) they are more often in contact with, with group contact potentially magnifying existing phonetic bias-driven asymmetries \citep{harrington_linking_2018}. Yet frequency of contact cannot be considered in a social vacuum. Work in the variationist sociolinguistic tradition (including \citealp{eckert_language_2000,labov2001principles,labov2007transmission,eckert2017phonetics}, etc. and associated formal/computational work such as \citealp{burnett_signalling_2019} and \citealp{kauhanen_replicatormutator_2020}) emphasizes the centrality of social evaluation in determining who imitates who and under what circumstances. An important idea in this line of work is that not only innovative groups, but also innovative \textit{speakers} are more likely to be imitated \citep{labov1990intersection,labov2001principles,harrington2017fronting,harrington_linking_2018}. Approaches stemming from accommodation theory, including `identity-projection' models and the notion of `style design' 
(for a review, see \citealp{auer2005role}), further emphasize that imitation is not automatic, but conditional on the social relationship between speaker and hearer. From the sociolinguistic perspective, then, understanding the dynamics of sound change requires understanding not just the source of a novel variant (which may indeed have come about due to some phonetic bias), but 
population-level variability in how variants are evaluated. 

Other work raises the possibility that the likelihood of propagation could involve evaluation of specific {variants} themselves. For example, \citet{baker2011variability} argue that adoption proceeds by imitation of variation, but that awareness of variation is limited to extreme targets. If learners only adopt variants they are aware of, and since extreme variants will by definition be rare, this helps explain why sound change itself is rare. \citet{garrett2013phonetic} consider the possibility that whether or not a given token (exemplar) is retained (stored) may be socially motivated. This could potentially mean one of two things: that certain learners are more prone to retain innovative variants, potentially for reasons outlined above, or that a given \textsc{variant} has some socioindexical value affecting the probability with which it will be stored or retained, independent of the speaker who utters it (or the hearer's relationship with/evaluation of that speaker). The idea that particular variants may be more salient to listeners is consistent with the sociolinguistic proposal that a variant must somehow be `marked' before it can take on valence (social meaning) \citep{hall-lew2021social,silverstein2003indexical}. 


The foregoing review, while far from comprehensive, highlights the range of external forces that may be involved in the actuation of sound change at the population level above and beyond phonetic biases. That is to say, while change might take place at the population level due to the enhancement of an asymmetric phonetic bias, we must also consider other situations characterized by asymmetries like frequency of interaction, social evaluation, or individual differences. 
This leads naturally to asking: \emph{what do the dynamics of actuation look like when invoking forces besides phonetic bias}?  There may well be substantive differences in the evolution of a variant which are dependent on our assumptions about how that variant is propagated: due to positive evaluation of a group, of particular individuals, or of the variant itself.  Because of the non-trivial mapping between individual learning/usage and population-level dynamics, computational modeling provides an attractive approach to understanding the potential contributions of different factors.

\subsection{Computational approaches to modeling propagation} 

Empirically evaluating the time course of sound changes usually requires access to phonetic data gathered over long timespans (whether from panel studies or using apparent-time data), such as the Philadelphia Neighborhood Corpus \citep{labov2013one}, the Atlas of North American English \citep{labov_atlas_2006}, the Queen's Christmas broadcasts \citep{harrington2000does}, or the Sounds of the City corpus (\citealp{stuart-smith2017changing,sonderegger2020structured,soskuthy2020voice}; see also Cox, Palethorpe, and Penney, this volume). Such datasets enable 
rich inferences about the dynamics of how changes unfold over time in a population setting. Computational modeling is a complementary approach, providing not only a way to study changes for which the necessary empirical data are lacking, but also a (relatively) quick way to test hypotheses and make empirical predictions. Moreover, a well-specified computational model provides a baseline against which to judge alternative explanations, and the implementation of a model -- the process of translating theoretical concepts into concrete parameters -- can prove invaluable in terms of evaluating and adjudicating between possible explanations, and generally helps to avoid the ``hazards of unaided reasoning''.\footnote{``With purely verbal arguments about evolutionary processes, it is too often the case that our conclusions do not follow from our assumptions. Unaided reasoning about the mass effects of many small forces operating over many generations has proven to be hazardous. Formalizing our arguments helps us understand which stories are possible explanations.'' \citep[][6]{mcelreath2007mathematical}} This is the case not only because intuitions based on purely verbal descriptions can often be misleading, but also because in complex settings like language change, it is far from obvious how different factors will interact: iterating weak forces over time generally leads to surprises.

A good example of this is \citet{baker2008addressing}, a computational exploration of the classical Neogrammarian hypothesis that sound change is fundamentally the accumulation of error. In his simulations, agents are connected in a social network based on relative prestige and interact only with other agents in their subnetwork. Tokens produced by each agent are subject to a small amount of random noise. Baker shows that such a model is unable to produce the typical s-shaped curve familiar from empirical studies of sound change, in which transition between periods of stability is initially slow but then proceeds quite rapidly. Baker’s model also underscores an important point: because most variation does \textit{not} result in change (what \citealp{kiparsky2015phonologization} calls the `non-phonologization problem'), 
any adequate model of sound (indeed language) change must be able to model stability, as well as change. As Baker aptly demonstrates, simple-minded implementations of phonetic bias are insufficient in this regard, since the introduction of the bias in such models inevitably leads to its adoption (cf. \citealp{pierrehumbert2001exemplar} and \citealp{wedel2004category}, who show something similar in terms of how unconstrained memorization of new exemplars leads to increasingly diffuse phonetic category distributions). 

Much like actuation may be initiated by forces other than production bias, stability may come about due to different mechanisms in different models. Several lines of modeling work have found that to meet the stability goal, it is 
necessary to include some kind of force promoting \textsc{contrast maintenance}, to keep separate phonetic categories stable, alongside an external force, such as a production bias, which induces change \citep{pierrehumbert2001exemplar,wedel2006exemplar,kirby2013role,kirby2015bias,soskuthy2015understanding}. Other work shows how individual- or group-level differences in evaluation can promote stability. 
For example, \citet{garrett2013phonetic} demonstrate how both stability and change can come about in a model where a listener or listener group disregards biased variants. These simulations 
are instructive in that they show how different results may arise from the same starting conditions under different assumptions about the nature of bias.


\subsubsection{Two modeling traditions: interactive-phonetic and population dynamics}

Every computational model must contain some simplifications, and different models contain different more/less articulated components depending on their goals. We focus here on two strands of the computational modeling of sound change literature which make different choices in these regards, but which share a focus on two central goals: (1) replicating the dynamics of change observed in real-world cases and in population settings, while (2) allowing for both stability and change in the face of bias factors.


The first literature can be exemplified by the `interactive-phonetic' agent-based model of Harrington and colleagues (hereafter IP-ABM: \citealp{harrington2017fronting,harrington_linking_2018,stevens2019associating,stevens2022individual,gubian2023phonetic}). While there are important differences between the various versions of this model, all implementations share some basic properties. First, they all consider interactions between a population of agents, with \textit{realistic} initial conditions, seeded with data taken from phonetic studies of real speakers. Second, the representation of individuals in the model is fairly complex, including increasingly sophisticated and well-articulated representations of phonetic and phonological categories and rules for how they are updated and change as agents interact. Third, such models assume finite population sizes and (usually) exclusively `horizontal' transmission in which all agents are part of a single generation that interacts with itself. The evolutionary dynamics of such models are inherently stochastic, meaning that typical behavior is obtained by running a simulation with a given set of inputs and parameterization many times. Finally, there is a common force which drives change in all of these models: the interaction between two distinct subpopulations, one of which shows the influence of a phonetic bias factor. 

Simulations in these papers explore how different initial configurations can lead to stability vs. change, and the results are then compared to the trajectories of sound changes in progress, some of them known historical changes. For example, \citet{stevens2022individual} consider the case of /s/-retraction, which is apparent in <str-> words in Australian English but not in Italian. In their IP-ABM simulations, Italian-initialized agents did not show evidence of /s/-retraction while English-initialized agents did. This demonstrates that both stability and change are possible within the IP-ABM framework (see also Jochim and Kleber, this volume).
At least in its current implementation, however, 
stability and change arise in the IP-ABM model not due to different values of model parameters, but as a function of the input data and initial conditions (how agents are initialized). Thus, while the internal representations of agents are complex, the exploration of dynamics is simplified. 

Another strand of this literature assumes a much simpler representation of each agent, and focuses on the study of the population-level dynamics as model parameters are varied. This includes approaches such as that of \citet{soskuthy2015understanding} as well as our own prior work \citep{kirby2013model,kirby2015bias}, described in more detail below, which builds upon earlier explorations of both sound change \citep{sonderegger2010combining,sonderegger2013variation} as well as syntactic change \citep{niyogi1996language,niyogi2009proper,niyogi2006computational}. 
These studies assume much simpler models of category representation and learning than the IP-ABM model, but emphasize the impact of 
\emph{varying} a range of model parameters, such as phonetic bias, categoricity bias, population structure, and variant frequency. Of central interest is mapping out the space of stable and unstable states as model parameters are varied (what Sóskuthy calls the `adaptive landscape'), which determine what path the population-level sound system will follow {for some given initial state}.  

A central claim of this latter literature is that (at least part of) the solution to the actuation problem lies in the \textit{population-level} dynamics: sound systems (characterized as a population-level distribution of how particular word or sound is pronounced) will always evolve towards stable states; thus, stability will (correctly) emerge as the default. A sound system only changes when a change in some model parameter makes the population's current state unstable. Like any complex adaptive system, such (phase) transitions tend to be {non-linear}: as a model parameter is changed (e.g., the degree of phonetic bias relative to categoricity bias) past a critical value, the population's current state suddenly becomes unstable (giving rise to the celebrated S-curve of linguistic change; see e.g., \citealp{blythe2012scurves}). In this general approach, the goal of the modeler is therefore to map out how changing model parameters changes what the stable states are.

If the approach typified by the IP-ABM model has as its goal to find model configurations which replicate known cases of change (but which do not erroneously predict change where none has occurred), models of the second type are assessed by their ability to meet three more abstract goals.
We seek models with a \textit{general} structure that we can show allows for (a) the possibility of stability in the face of bias; (b) the possibility of change in the face of bias; and 
(c) a nonlinear transition from stable variation to change as a function of system parameters. 
 
\subsection{The present study}

In previous work, we have described a modeling framework (outlined in §2 below) in which categoricity bias can enforce stability even in the face of production bias, and in which production bias can induce change even in the presence of categoricity bias. Such a model is \textit{adequate} in the sense that it allows for both stability and change. However, as discussed in §1, production biases are not the only forces which can perturb a population from equilibrium. This leads us to pose the following research questions: 

\begin{enumerate}
    \item \emph{Does the use of production bias as a perturbing force have a {unique} dynamics}, or can a nonlinear transition from one stable state to another emerge when a different force is employed? 
    \item  If it can, will \emph{any kind of external force produce the same dynamics at the population level}, or do different forces have different dynamics?
 \end{enumerate}


Note that in this setting, `dynamics' means the deterministic dynamics of well-mixed and infinitely large populations which define a landscape of \textit{possible} trajectories of change, rather than the properties of specific temporal trajectories in simulations corresponding to a real-world instance of a particular sound change (as is typically the focus in agent-based simulations such as the IP-ABM). 

We consider models containing two external forces beyond phonetic production bias: (1) contact between subpopulations with different stable speech patterns, and (2) the weight agents give to tokens depending on their source (based either on the speaker's group membership or individual identity) or the extent to which a token is phonetically innovative.  We assess each model based on whether both stability and change are possible as model parameters (= the strength of the external force) are varied. In this paper, we focus on the particular example of the propagation of a phonologized coarticulatory variant, but the general framework is applicable to many types of change involving a wide range of non-phonetic production biases.

The broader question addressed by this exercise is: can we safely assume that \textit{any} proposed force driving change \textit{could} lead to change, iterated over time in a population?  This fundamental assumption is implicit in a great deal of theorizing about sound change, but remains untested. The answer is not obvious, because (as we will see) even in simple models where individual agents are anything but sophisticated, 
unintuitive outcomes may still result at the population level (see also Jochim and Kleber, this volume). If the answer turns out to be `no', this would mean that a thorough understanding of population dynamics, in addition to their source, will be necessary to get to grips with how particular changes actuate.





\section{Modeling framework\label{sec:framework}}

We begin by reviewing the simulation framework introduced in \citet{kirby2013model,kirby2015bias}, which forms the basis for the simulations we present here.\footnote{Code for simulations reported in this paper can be found at  \url{https://osf.io/b7sgq/}. Code for simulations reported in \citet{kirby2015bias} can be found at \url{https://github.com/kirbyj/evomod/tree/master/ks}.} Our notation follows  \citet{kirby2015bias}, which can be consulted for more detail on the general framework.
In that work, as well as in the simulations discussed in the present paper, we use the phenomenon of West Germanic primary umlaut, a textbook example of the phonologization of coarticulation, as an illustrative example (Table \ref{tab:umlaut}).  In pre-Old High German (OHG), short low /a/ was fronted and raised to /e/ when a high front vowel or glide occurred in the following syllable, as in *[gasti] $>$ /gesti/ (modern German \textit{Gäste}). At some stage, the conditioning vowel was weakened, but presumably not before the raising of the stem vowel was firmly established. We focus therefore on the state of affairs which gave rise to the shift from proto-West Germanic to pre-OHG, i.e., the establishment of an [e]-like allophone of /a/ in the context of a following /i/.\\

\begin{table}[ht]
\caption{Primary umlaut in West Germanic (after \citealp[71]{iverson1996primacy}).\label{tab:umlaut}}
\centerline{
\fbox{
\begin{tabular}{llll}
\it{WGmc} & \it{Pre-OHG} & \it{OHG} & \textit{Mod. German}\\[2pt]
\midrule
{*}gasti & gesti & gest & \textit{G\"{a}ste}\\
{*}lambir & lembir & lemb & \textit{L\"{a}mme}\\
{*}fasti & festi & fest & \textit{fest}\\
\end{tabular}}
}
\vskip 0.15in
\end{table}

In \citet{kirby2015bias}, we began by making as few assumptions as possible and built up an increasingly complex model until we had an \textit{adequate} account of actuation, defined as a model which generates the following regimes:

\begin{enumerate}
\item[(a)] the stability of \textit{limited} coarticulation in the population, as in pre-Old High German;
\item[(b)] the stability of \textit{full} coarticulation in the population, as in Old High German; and 
\item[(c)] sudden and nonlinear \textit{change} from stable limited to stable full coarticulation, as model parameters are varied.
\end{enumerate}

We refer to (a)-(c) as our \textsc{modeling goals}.

\subsection{Linguistic setting}

Agents in our model are represented by a simple lexicon consisting of three vowel categories \{{V}$_1$, {V}$_2$, {V}$_{12}$\}, where {V}$_{12}$ represents V$_1$ in the coarticulation-inducing context of V$_2$ (e.g., /a\underline{\hskip 0.1in}i/, as opposed to plain /a/ or /i/)\footnote{We employ this notation, rather than using e.g. /e/, to underscore the fact that \textit{phonemically} we are modeling a state where there are two categories: that is, /a\underline{\hskip 0.1in}i/ is a (stable) variant of /a/, rather than a separate phoneme.}. Vowels are represented as first formant (F1) values. For simplicity, the F1 distributions of V$_1$ and V$_2$ are assumed to be normal ($V_1 \sim N(\mu_a, \sigma_a^2)$, $V_2 \sim N(\mu_i, \sigma_i^2)$), known to all learners, the same for all learners, and stable over time.

The F1 distribution of V$_{12}$ is normal, with fixed variance as for V$_1$ and a mean we denote by $c$:
\begin{equation}
V_{12} \sim N(c, \sigma_a^2)
\label{eq1}
\end{equation}
We will sometimes refer to V$_{12}$ (or equivalently, $c$, which determines the distribution of V$_{12}$) as the \emph{contextual variant}.  $c$ lies between the means of V$_1$ and V$_2$, and $\mu_a - c$ is the extent to which /a/ (V$_1$) is coarticulated in the context of /i/ (V$_2$).

\begin{figure}[!ht]
\begin{center}
\includegraphics[trim=0 10 0 50,clip]{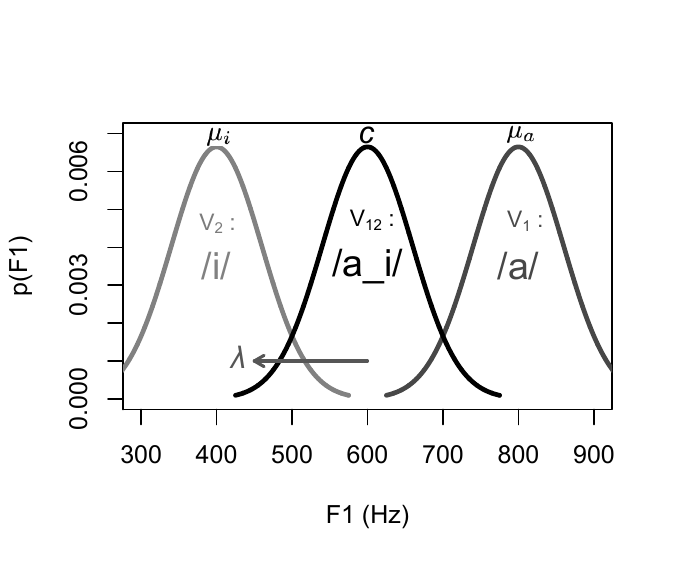}
\caption{Lexicon and learning parameters. $\mu_a$ and $\mu_i$ are the means of normally-distributed vowel categories V$_1$ (/a/) and V$_2$ (/i/).   $c$ is the mean of the normally-distributed contextual variant, V$_{12}$. $\lambda$ represents the strength of the bias favouring coarticulated variants. \label{fig:llp}}
\end{center}
\end{figure}

In addition, we assume that productions of $V_{12}$ may be subject to phonetic production bias, represented by a quantity $\lambda$ which describes the tendency of a speaker to over- or undershoot articulatory targets.\footnote{In \citet{kirby2015bias}, the production bias is allowed to vary  normally across tokens. For simplicity we assume a single fixed production bias here.}  Thus, the actual productions of an agent with contextual variant $c$ follow the distribution:
\begin{equation}
F1 \sim N(c - \lambda, \sigma_a^2)
\end{equation}
(Note that the bias adjusts the mean of V$_{12}$ by negative $\lambda$, because V$_2$ has lower F1 than V$_1$.) This scenario is illustrated in Fig.~\ref{fig:llp}.

\subsection{Learning and evolution}

We assume that agents are divided into discrete generations, each containing a very large number of agents.\footnote{In the limit of infinitely-large populations, the evolution of the distribution of $c$ becomes deterministic, and can be analyzed as a dynamical system (see e.g., \citealp[][chapter 5]{niyogi2006computational}).} Learners in generation $t+1$ receive $n$ examples of $V_{12}$ drawn from generation $t$. Each learner's task is to infer $c$.  We assume that learners apply a learning algorithm which is `rational', in the sense that they assume each token in their learning data to be generated according to Equation~\ref{eq1}, and estimate the most probable value of $c$.  That is, each learner's task is simply to infer how much  /a/ is produced like /i/ in the context of /i/ (/a\underline{\hskip 0.1in}i/). 

\begin{figure}[!ht]
\begin{center}
\includegraphics{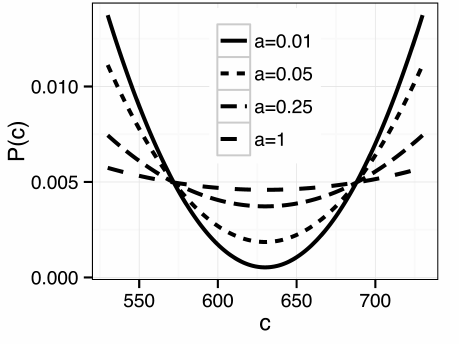}
\caption{Prior distribution over $c$, for values between  the means of V$_2$ and V$_1$ ($\mu_i = 530$, $\mu_a = 730$). The parameter $a$ controls the strength of the categoricity bias, with values nearer to 0 corresponding to a greater preference for values of $c$ near either endpoint.\label{fig:cb}}
\end{center}
\end{figure}

In \citet{kirby2015bias}, we allowed this basic model to vary in two ways. 
The first was changing the learning algorithm agents applied, by adding a \textsc{categoricity bias}: the degree to which agents preferred /a\underline{\hskip 0.1in}i/ tokens to conform to the fixed distributions of /a/ and /i/. This was implemented as a prior over the distribution of $c$. We explored two cases, a simple Gaussian prior where $c$ was biased to be near $\mu_a$ (corresponding to a preference/expectation for /a\underline{\hskip 0.1in}i/ to be realized as /a/) and a more complex polynomial prior representing a bias for $c$ to be a value around either $\mu_a$ or $\mu_i$ (corresponding to either no coarticulation, or full coarticulation). The strength of this `complex prior' is controlled by a parameter $a$, as shown in Fig.~\ref{fig:cb}.  All simulations in the current paper use this prior to capture categoricity bias.

\begin{figure}[!ht]
\begin{center}
\includegraphics[width=1.0\linewidth]{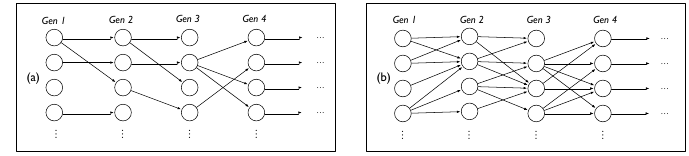}
\caption{Two types of population structure considered in models in \citet{kirby2013model,kirby2015bias}: (a) Single-teacher scenario, where each learner in generation $t+1$ receives all her data from a single teacher in generation $t$. (b) Multiple-teacher scenario. Each data point comes from a random teacher, each chosen uniformly at random (with replacement) from teachers in generation $t$. \label{fig:ps}}
\end{center}
\end{figure}

The second issue we addressed was that of population structure: whether agents were assumed to learn from single teachers, a frequent assumption in computational modeling of language change (often called `iterated learning': \citealp{smith2009iterated,griffiths2007language}, \citealp[S.][]{kirby2007innateness}), 
or multiple teachers, as illustrated in Fig.~\ref{fig:ps}. Because we observed significant differences between the dynamics in single- vs. multiple-teacher settings \citep[c.f.][]{niyogi2009proper}, and because only the results from the multiple-teacher settings were consistent with the extant sociolinguistic evidence, we restrict our attention to the multiple-teacher setting in the present paper.\footnote{\citet{kirby2015bias} also considers the case where each agent learns from two (randomly chosen) teachers, which we do not consider here because the dynamics were similar to the multiple-teacher case.}

Considering the ensemble of all agents in generation $t$, the state of the population at $t$ can be characterized by a {probability distribution} describing how likely different values of $c$ are. Formally, this is the distribution of a random variable, which we write as $C^t$. Similarly, the values of $c$ learned by agents in generation $t + 1$ can be characterized by $C^{t+1}$, whose probability distribution describes how likely different values of $c$ now are. For simplicity, we assume that the number of agents per generation is very large. The evolution of the distribution of $c$ is then deterministic, making its behavior more easily analyzed as a dynamical system. This and several other aspects of our modeling framework, such as the assumption that generations are discrete, are shared with the broader literature on dynamical systems models of language change (e.g., \citealp{niyogi1996language,niyogi2006computational,yang_internal_2000}; see §1.3.1).

\subsection{Assessment}

In this framework, given a choice of learning algorithm, population structure, and degree of phonetic bias, we may characterize the evolution of the distribution of $c$ and determine the extent to which it satisfies our modeling goals. When learners are assumed to have a categoricity bias like that in Fig.~\ref{fig:cb}, analytic solutions become intractable, and we must proceed by simulation to determine how the distribution of $c$ evolves. For practical reasons, in this chapter we present simulations over a limited number of generations (between 50 and 500), but it is important to keep in mind that the stable distribution of $c$ remains a fundamentally deterministic, rather than stochastic, function of the parameter settings. As a result, the number of generations our simulations require to converge to a stable state is not directly comparable to simulation time in stochastic agent-based simulations such as \citet{harrington2017fronting}, \citet{stevens2019associating}, or Jochim and Kleber (this volume). Our goal is less to characterize the speed, slope, or trajectory of change than to determine the parameter regimes in which change occurs at all, as well as which change will in a occur in a particular regime.

\begin{figure}[!ht]
\begin{center}
\includegraphics[width=0.45\linewidth]{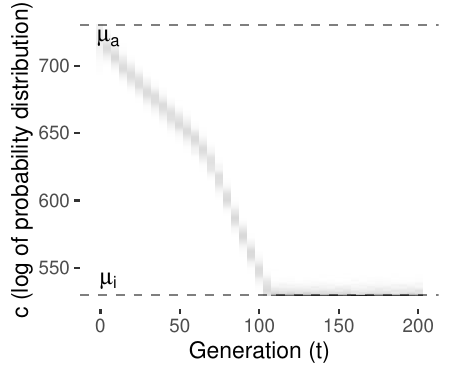}
\includegraphics[width=0.45\linewidth]{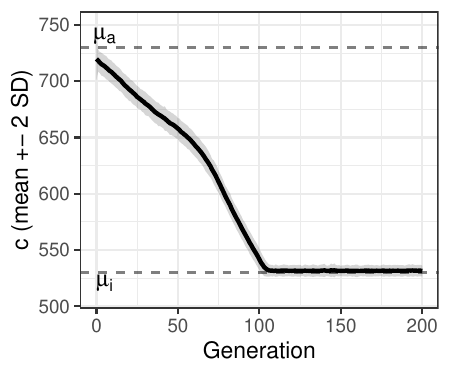}
\caption{Example of the distribution of $c$ in the population at time $t$ (probability density function $P(C^t)$). The population starts with minimal coarticulation at $t=0$ ($c \sim N(720, 10^2)$) and ends with full coarticulation by $t=100$.  All agents have weak categoricity bias ($a=0.02$) and strong production bias ($\lambda = 2$). Left panel: shading is proportional to $P(C^t)$. Right panel: solid line and shading show the mean $\pm$ 2 standard deviations of $P(C^t)$.\label{fig:patt}}
\end{center}
\end{figure}

\begin{figure}[!ht]
\begin{center}
\includegraphics[scale=0.35]{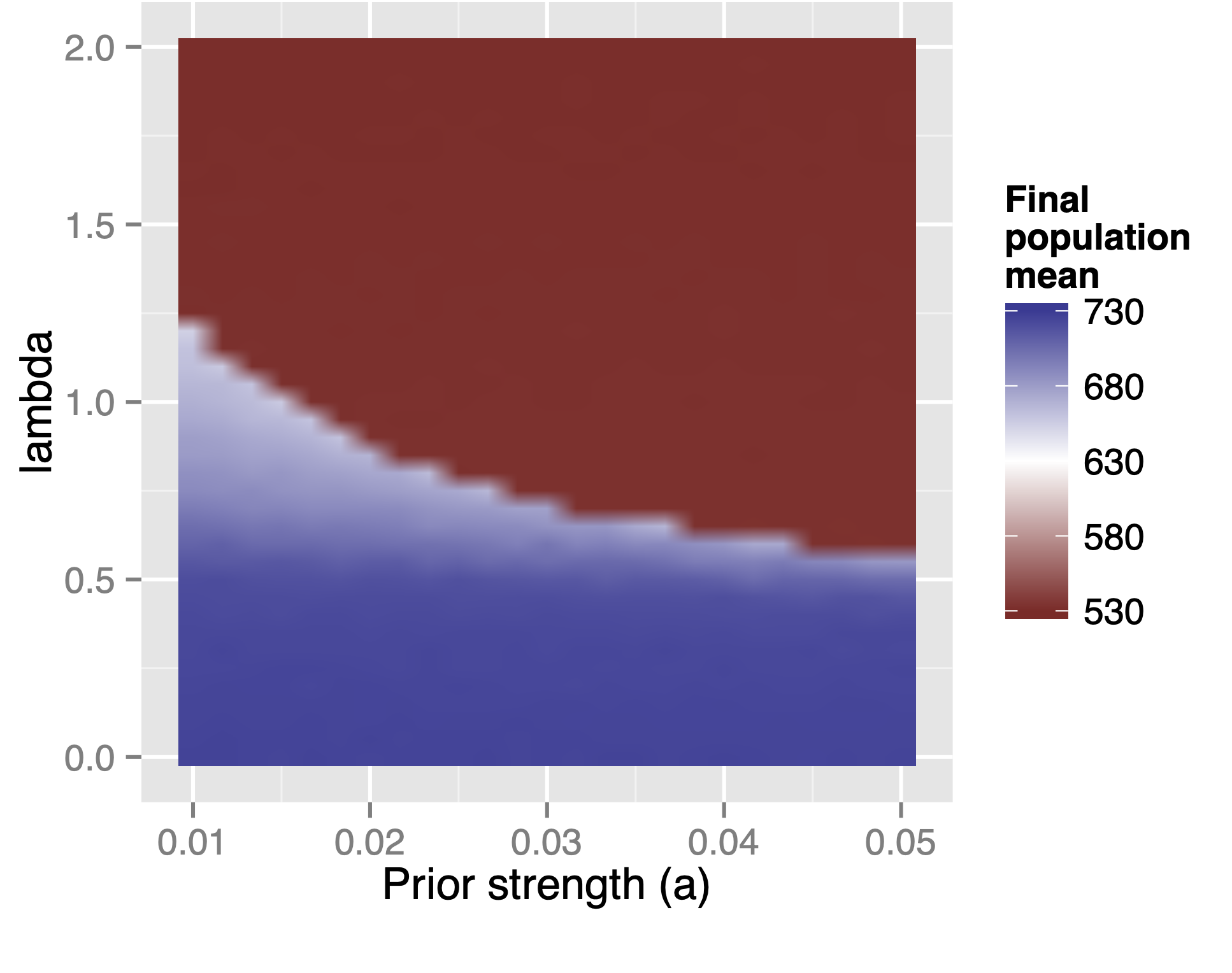}
\caption{Mean value of $c$ in the population in its stable state, starting from a population of agents with a minimally coarticulated /a\underline{\hskip 0.1in}i/ variant ($c \sim N(\mu_a - 10, 10^2)$) with production bias $\lambda$ and categoricity bias $a$ (smaller $a$ = stronger bias).  The final mean changes non-linearly as $\lambda$ and $a$ are varied. Red and dark blue correspond to full/no coarticulation of $V_{12}$, respectively. 
\label{fig:ks13}}
\end{center}
\end{figure}

Fig.~\ref{fig:patt} shows an example for the setting where all agents have a minimally coarticulated variant in generation 1 ($c$ is normally distributed with mean $\mu_a - 10$), and there some production bias ($\lambda = $ 2) along with weak categoricity bias ($a = $ 0.02). The results can be visualized in two ways. The left panel shows the evolution of the \emph{entire distribution} of $c$, while the right panel summarizes this distribution by showing just the evolution of its mean (representing the `average pronunciation' of V$_{12}$). From either representation, it is clear that $P(C^t)$ quickly converges to having most mass around 530 Hz (= $\mu_i$, the F1 of V$_2$), meaning that the population converges on a state of \emph{full} coarticulation.\footnote{Settings of other parameters for this simulation and those shown in Fig.~\ref{fig:ks13}: $\mu_a = 530$, $\mu_i = 730$, $\sigma_a = \sigma_i = 50$, $M = 500$, $n=100$.}



In \citet{kirby2013model,kirby2015bias}, it was shown that only a model with \textit{both} production and categoricity biases could achieve all three modeling goals (stable limited coarticulation, stable full coarticulation, and change from one to the other).  When the degree of categoricity bias is held constant, increasing $\lambda$ past a critical value causes a rapid transition from stable limited coarticulation (low $\lambda$) to stable full coarticulation (high $\lambda$).  This dynamics of this case are characterized by a trade-off between the strength of categoricity bias and production bias, with a sudden change from limited to full coarticulation. This `adaptive landscape' is illustrated in Fig.~\ref{fig:ks13}, which shows the degree of coarticulation the population will end up with, as $a$ and $\lambda$ are varied, for the same starting state (limited coarticulation), in a concrete example. To generate this figure, we ran a simulation like the one shown in Fig.~\ref{fig:patt} using a range of values of $a$ and $\lambda$, and recorded where the average degree of coarticulation stabilizes after a large number of generations ($t = 2500$). In Fig.~\ref{fig:patt}, this stable state is near $\mu_i$ (= 530 Hz).


Before moving on to our extensions, we emphasize two further aspects of this earlier work. First, in models with both categoricity bias and production bias, 
phonologization is not inevitable (cf. \citealp{baker2008addressing}): the mere presence of a production bias ($\lambda > 0$) does not entail that a novel variant will inevitably become the dominant variant in the speech community. Second, the dynamics of phonologization in this model are non-linear: a very small change in the degree of bias can result in phonologization (sudden population-level change) or not (population-level stability), simply depending on the state of the system (e.g., the amount of production vs.\ categoricity bias). In particular, there is a clear bifurcation: once $\lambda$ is large enough relative to $a$, the stable state will be one of full coarticulation. It is thus not necessary to postulate any additional mechanism to explain both phonologization and `non-phonologization'.

\section{Extensions: contact and social weight}

In this section, we extend this framework to consider two additional types of external forces beyond production bias: \textsc{contact} between subpopulations (§\ref{sec:model1}) and \textsc{social weight}\footnote{What we are referring to as social weight encompasses the more traditional concept of `prestige', but as discussed in \citealp{salmons2021sound} (Ch. 7), this term is typically not well-defined, so we explicitly avoid it here.} at the level of variants, individual speakers, and groups (§\ref{sec:models2-4}).  We assess each model based on whether both stability and change are possible as model parameters are varied.  
Our first extension considers a case conceptually similar to that explored by \citet{harrington2017fronting,stevens2019associating,gubian2023phonetic} \textit{inter alia}, in which a bias variant is firmly established in the speech of one group but not another. In the context of our modeling framework, the question we ask here is: are both stability and change possible when heterogeneous groups interact?

\subsection{Model 1: Subpopulations in contact\label{sec:model1}}

\begin{figure}[!ht]
  \begin{center}
    \includegraphics{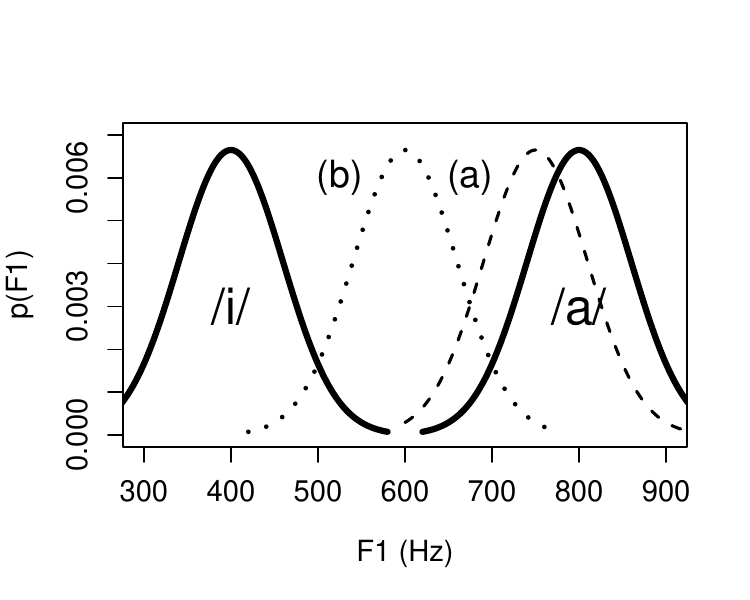}
    \caption{Lexical distribution of V$_1$, V$_2$, and $c$ at the outset of the subpopulation mixing simulations. Distribution (a) (dashed line) is the distribution of $c$ for group A; distribution (b) (dotted line) is the distribution for group B.}
    \label{fig:subpopSetup}
\end{center}
\end{figure}

In this setting we consider a population consisting of two groups. Group A is characterized by a contextual variant exhibiting little or no coarticulation, while for Group B, this same variant is extremely coarticulated (Fig.~ \ref{fig:subpopSetup}). Let $aProb$ be the probability that a Group B agent learns from (equivalently, imitates/stores an exemplar from) a Group A agent, and let $bProb$ be the probability a Group A agent learns from a Group B agent. In this setting, there is no production bias ($\lambda=0$), and the $a$ parameter controlling the strength of the categoricity bias was set at 0.01, encoding a strong dispreference for intermediate variants (see Fig.~\ref{fig:cb}). 
In other respects, the simulation procedure was the same as outlined in §\ref{sec:framework}.

\begin{figure}[!ht]
  \begin{center}
    \includegraphics[width=\textwidth]{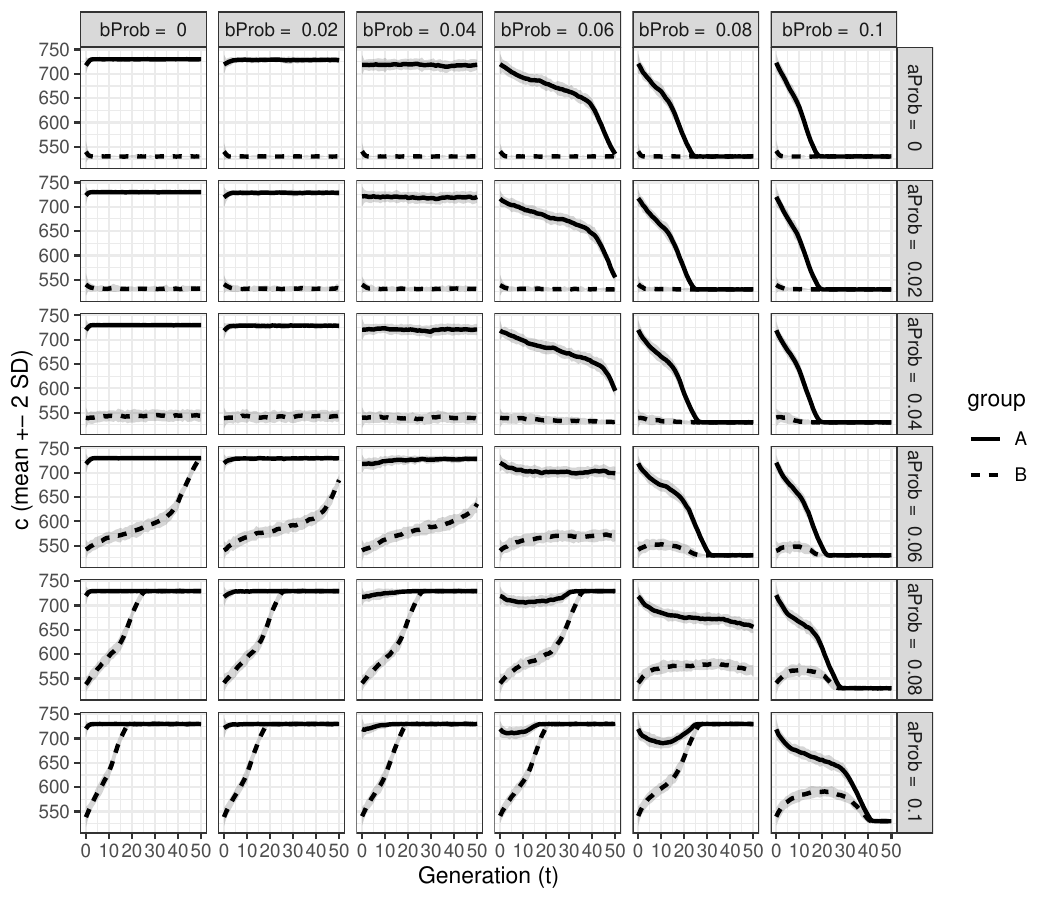}
    \caption{Results of subpopulation simulation modeling contact between groups with minimal coarticulation (Group A) and full coarticulation (Group B) at $t=0$.}
    \label{fig:subpopResults}
\end{center}
\end{figure}

A selection of the results of these simulations after 50 simulation epochs are given in Fig.~\ref{fig:subpopResults}, which shows the evolution of $c$ for each group. The main result is that the starting mean F1 value $c$ characterizing each group's stable state (Fig.~\ref{fig:subpopSetup}) appears to be stable even when there is some degree of interaction between them (upper left panels). 
However, obtaining (and storing/imitating) just 5\% of training examples from a different group can be enough to induce the \textit{entire} population to converge to one or the other group's mean. There may also exist regimes in which both groups converge on an intermediate distribution, i.e., a more extreme version of the $aProb = bProb = 0.06$ panel shown in Fig.~\ref{fig:subpopResults}. 

In regimes in which there is a sizable asymmetry between $aProb$ and $bProb$, the direction of convergence appears to be predictable. The panels in the lower right quadrant of  Fig.~\ref{fig:subpopResults} in which $aProb = bProb$, particularly the case where $aProb=bProb=0.1$, merit additional mention. That both groups are seen converging to the group B mean in the lower right panel of Fig.~\ref{fig:subpopResults} is unlikely to consistently replicate; repeating this simulation many times, with the same parameter settings, should yield approximately as many outcomes which converge to the group A mean. This is because simulation requires that we work with a finite population sample, whereas the `true' (deterministic) stable state is contingent on the assumption of infinite populations. 


This simplistic scenario could be further complicated in many ways, e.g., by introducing differences in the orientations of the distributions between groups. However, it already shows us that, in the context of our modeling framework, the subpopulations in contact scenario fulfills our modeling goals, with a dynamics very similar to that seen by the introduction of a production bias: there exist parameter regimes in which both stability and change are possible. 

\subsection{Models 2-4: social weighting\label{sec:models2-4}}

The next three models explore whether both stability and change are possible in the presence of social value associated with more coarticulated \textit{variants}, 
 with valued \textit{speakers} who coarticulate more, and with \textit{groups} characterized by coarticulation. We are intentionally remaining agnostic as to exactly what motivates the association of `social value' with a variant, a speaker, or a group, but see Section \ref{sec:intro}, as well as \citet{garrett2013phonetic} and \citet[][chapter 7]{salmons2021sound}, for some discussion. 


In our social weighting settings, each token $y_i$ is transmitted as a vector of an F1 value together with a \textsc{social weight} value $w_i \in [1, w_{max}]$. How this value is determined varies across the three models. In Model 2, higher social weight is associated with more coarticulated \textit{tokens}, i.e., tokens of /a\underline{\hskip 0.1in}i/ are more highly weighted the more similar they are to /i/. In Model 3, tokens of /a\underline{\hskip 0.1in}i/ from a high-coarticulation \textit{group} are given greater weight, so exposure to highly coarticulated tokens of /a\underline{\hskip 0.1in}i/ is modulated by probability of group interaction. And in Model 4, the weight is associated with tokens of /a\underline{\hskip 0.1in}i/ from individual \textit{teachers} who coarticulate more. These weights are distinct from the probabilities (\textit{aProb} and \textit{bProb}) with which an agent of one group learns a token from another.

As before, the learner's task is to estimate $c$ using the weighted average of $y_i$. In these scenarios, however, tokens with values of $w_1>1$ (i.e., those that are either more coarticulated, or from teachers who  coarticulate more), will have greater influence on the evolution of $c$ in the population. 

\subsubsection{Model 2: social weighting by variant\label{sec:model2}}

\begin{figure}[!ht]
  \begin{center}
    \includegraphics[width=0.75\textwidth]{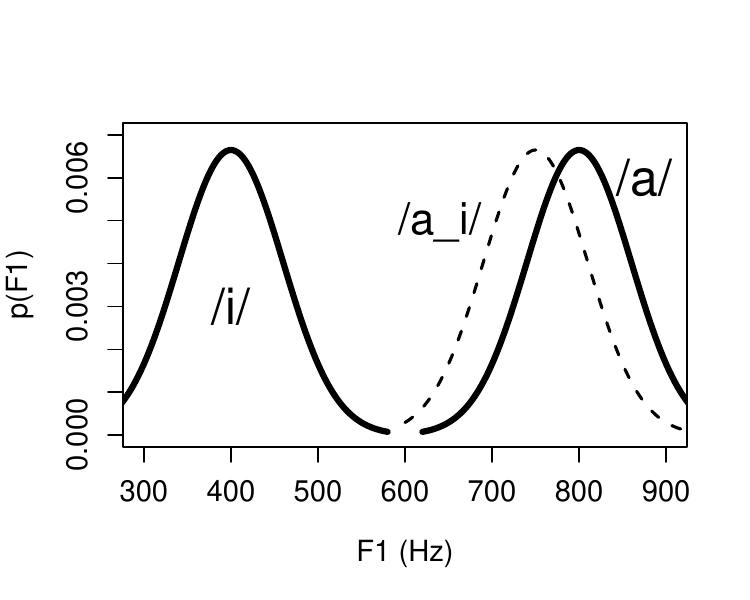}
    \caption{Initial distribution of  parameter $c$ (dashed line: mean pronunciation of V$_{12}$) in the Model 2 population at time $t=0$.\label{fig:sw1setup}}
\end{center}
\end{figure}

In the Model 2 simulations, we return to a setting with a single population in which /a/ is only slightly coarticulated in the context of /i/ (Fig.~\ref{fig:sw1setup}). In these simulations, we vary the parameter $w$ controlling the reference social weight accorded to coarticulated variants. The stronger the weight, the more that more strongly coarticulated tokens of $y_i$ (i.e, those produced with a lower F1, more like /i/ than /a/) contribute to the maximum likelihood estimate of the /a\underline{\hskip 0.1in}i/ category. 

\begin{figure}[!ht]
  \begin{center}
    \includegraphics{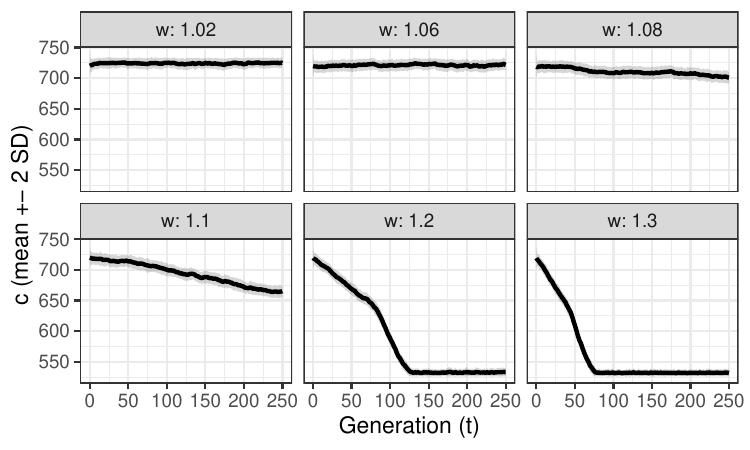}
    \caption{Model 2 simulation: Distribution of $c$ in the population at time $t$ ($C^{t}$) for varying weights $w$ indicating preference for the coarticulated variant.\label{fig:sw1}}
\end{center}
\end{figure}

The results of simulations for different values of $w$ run for 250 generations are shown in Fig.~\ref{fig:sw1}. Stability can be preserved when coarticulated variants have some social weight, but  having the social weight of the coarticulated variant be just 10~\% more than that of the uncoarticulated variant can be enough to induce change to full coarticulation in the whole population. Note that in this scenario, it is assumed that all learners weight coarticulated variants the same and that this weight never changes over time; as such, even for low $w$, convergence to the coarticulated variant may be inevitable given enough simulation generations. Further exploration of the parameter space would be required to determine this for certain. Once again, however, we see that (a) the introduction of a bias does not make change inevitable, and (b) the dynamics of the change are rapid and non-linear for at least some parameter settings -- where by `rapid and non-linear' we mean that small numerical changes in parameter settings in certain regions of the parameter space can result in qualitatively different outcomes.

\subsubsection{Model 3: social weighting by group\label{sec:model3}}

Model 3 has the same basic architecture as Model 1 (§\ref{sec:model1}), but with two additional parameters $aWeight$ and $bWeight$, corresponding to how much data from group A is weighted for a learner in group B and how much data from group B is weighted for a learner in group A, respectively. Weight values ranged from 0 to 1. As in Model 1, learners in each generation learn from teachers both in their own group and, potentially, the other group. In Model 1, we saw how for low settings of $aProb$ and $bProb$ (the parameters controlling the number of tokens learners receive from the opposing group), the distribution of $c$ in both groups (one with no coarticulation and one with stable coarticulation) would remain constant. In Model 3 we experimented with different values of these parameters, as well as with the weights given to tokens from each group.

\begin{figure}[!ht]
  \begin{center}
    \includegraphics[width=1\textwidth]{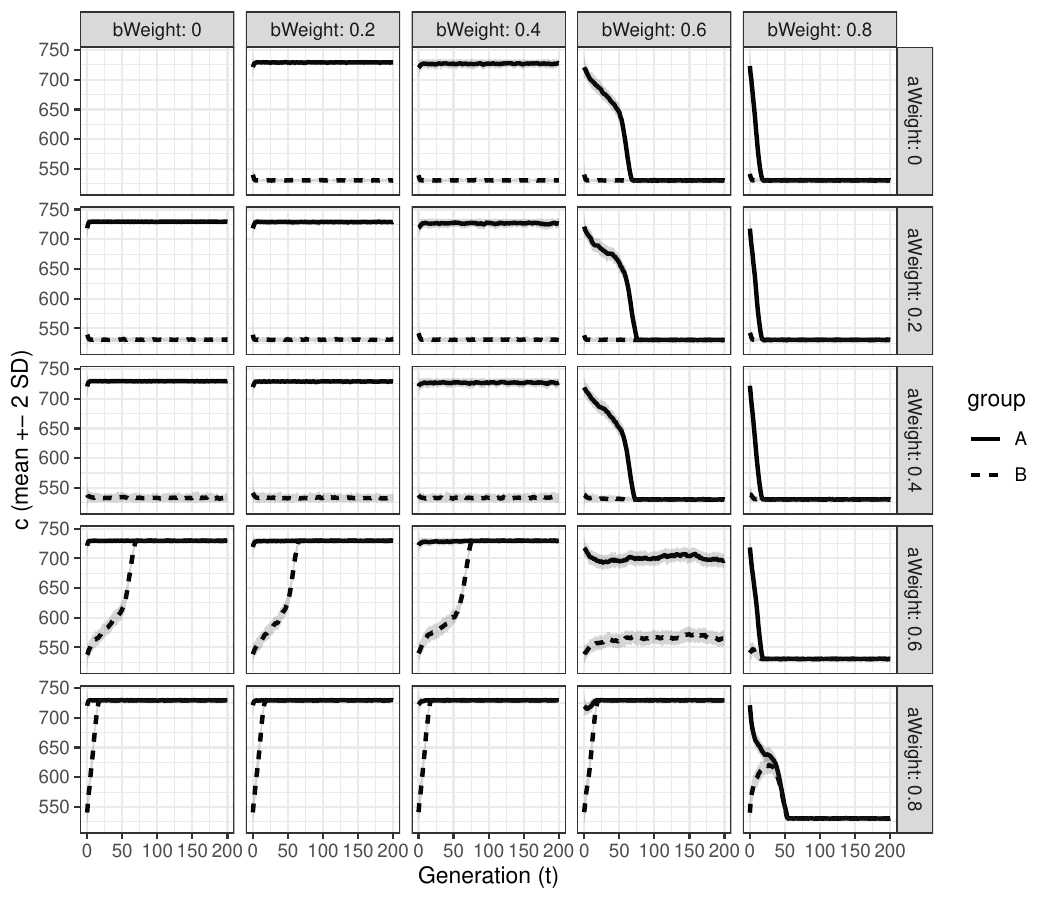}
    \caption{Results of Model 3 simulations for fixed $aProb=bProb=0.3$ and varying weights $aWeight$ and $bWeight$. \label{fig:model3}}
\end{center}
\end{figure}

While space does not permit a visual display of the entire parameter space exploration, some representative results are shown in Fig.~\ref{fig:model3} for the case where $aProb = bProb$ (using the value $aProb = bProb = 0.03$).  
Again, here, both stability and change are possible. Stability can be preserved even when tokens from the high-coarticulation group B are socially valued by low-coarticulation group A ($bWeight=0.2-0.5$). But even a small preference for tokens from the coarticulating group 
can be enough to induce change to full coarticulation in the whole population. Simulation results for other low values of $aProb$/$bProb$  (e.g., $aProb = bProb$ 0.05) look qualitatively similar, though the exact values of $aWeight$/$bWeight$ where transitions from stability to change occur differ. 

\subsubsection{Model 4: social weight by individual\label{sec:model4}}

Finally, in Model 4 we return to a single population, and consider the case where social weight values are associated with individual speakers. In this scenario, we associate every teacher $m \in M$ in generation $t$ with a social weight value $w_m$ as well as a value $c_m$, the mean of their coarticulated variant. If these happen to be positively correlated -- i.e., if tokens from teachers who coarticulate more also have higher social weight -- this implies more coarticulation in generation $t+1$, which \textit{could}, but need not, accumulate and lead to change (cf.~\citealp{baker2011variability}). The degree of correlation between the social weight $w_m$ of any individual teacher $m$ and their coarticulation parameter $c_m$ is controlled by a simulation-level parameter $\rho$, ranging from 0 (uncorrelated) to 1 (perfect correlation, so the strongest coarticulators also have the highest social weight). Individual teacher weights were sampled at random from the range $\{1, w_{max}\}$.\footnote{The exception to this was for speakers in the first 25 generations, for whom weights were assigned as the weighted average of the teacher's coarticulation parameter $c$ and the uniformly randomly sampled value in $\{1, w_{max}\}$. This was done to encourage individual simulation runs to start in roughly similar areas of the parameter space.}


\begin{figure}[!ht]
  \begin{center}
    \includegraphics[width=1\textwidth]{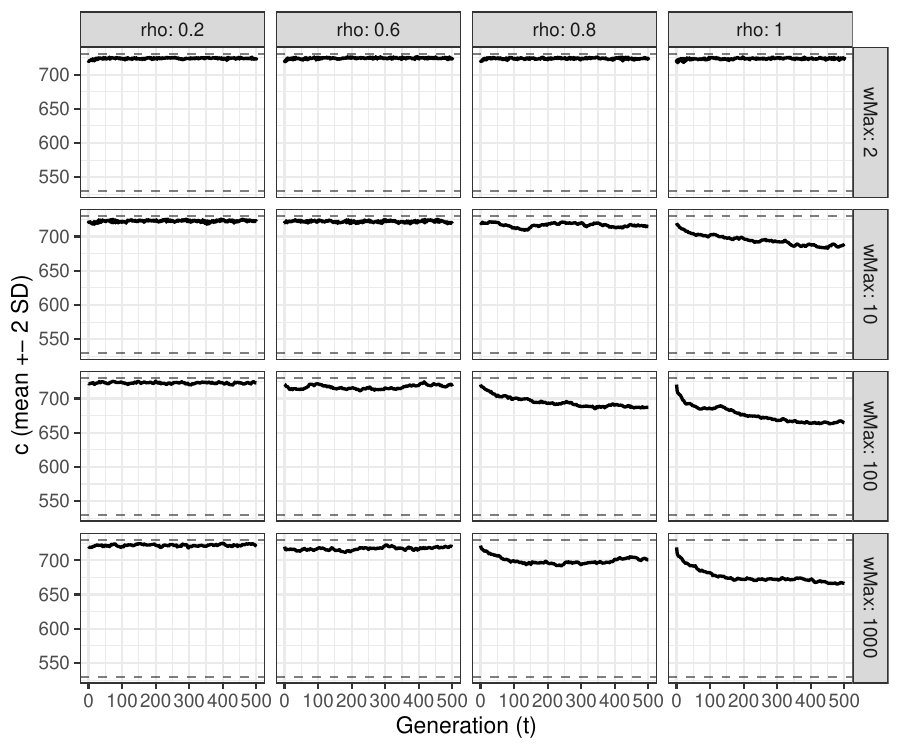}
    \caption{Results of Model 4 simulations for varying maximum weight $w_{max}$ and correlation $\rho$ between $w$ and $c$ for any individual teacher $m$. \label{fig:model4}}
\end{center}
\end{figure}


The results for several values of $\rho$ and $w_{max}$ are given in Fig.~\ref{fig:model4} (for runs of 500 generations). In this setting, stability is the default; it is only when $\rho \approx 1$ that we observe trajectories that appear to indicate change. 
In other words, in this setting, change requires a near-perfect correlation between coarticulation and social weight, where (tokens from) individuals who coarticulate are weighted 100-1000 times higher than those who do not. 
These findings are arguably expected given the way that weights are assigned to teachers (uniformly at random) and how learners receive input from teachers (also uniformly at random) in this setting. However, further simulations with many more generations are needed before any firm conclusions can be drawn.

\section{General discussion}

In this paper, we have considered the question of whether, in a population setting with a categoricity bias, propagation of a phonetic production bias has a \textit{unique} dynamics (question 1), and if not, will \textit{any} kind of driving force produce the same dynamics (question 2). To study these questions, we compared our earlier work with two new external forces: interacting groups with different stable states (Model 1, §\ref{sec:model1}), and various types of social weights assigned to variants, groups, or individuals (Models 2-4, §\ref{sec:models2-4}). 

The dynamics of Model 1, in which subpopulations with different distributions of a bias variant were put in contact, were broadly similar to those observed in \citet{kirby2015bias}: there exist parameter setting regimes (here, the degree of contact between groups) in which both stability and change are possible. 
Similar dynamics were observed in Models 2 and 3 (§\ref{sec:model2} and \ref{sec:model3}), in which social weights were assigned to variants and groups, respectively. Thus the answer to our first question -- does the introduction of a phonetic production bias have a \textit{unique} dynamics? -- is clearly `no'. This was not a foregone conclusion, given that previous models including phonetic production biases have had extremely idealized or simplified population structures.

Models 2, 3, and 4 are all implementations of social weight. Yet the dynamics of Model 4 (§\ref{sec:model4}) were markedly different from the others, in the sense that there is less of a qualitative difference between the regimes compared with the previous models. Thus the answer to our second question -- will \textit{any} kind of driving force produce the same dynamics? -- is also `no'. 

Why are the dynamics of Model 4 so different? The answer has less to do with how weights are assigned than it does with the correlation between weights and observations. In Model 2, where social weight was associated with {variants}, the correlation between weights and observations is perfect: the more strongly an observation was coarticulated, the more highly it was weighted, by all learners across all generations. In Model 3, the social weight is associated with groups, but applies to all observations equally. That is, an /a/-like token of /a\underline{\hskip 0.1in}i/ from the highly weighted group is valued just as much as an /i/-like token. The correlation between weights and observations is therefore somewhat reduced. In Model 4, where weights are properties of individuals, the correlation is weakened even further. For the scenarios we explored, in which the distribution of $w$ in the population was uniform, near-perfect correlation between weights and observations was required for change to be seen in the population-level distribution of $c$.

The results of our simulations cannot be taken as evidence that any of the scenarios we have considered -- production bias, interacting subpopulations, or differences in social weighting -- are responsible for any particular empirically observed instance of sound change. Given the current state of our scientific understanding of sound change, it is far from clear how the parameter values in any computational model should be set on the basis of real-world properties of utterances, individuals, and social groups. This is at least in part a consequence of our modeling strategy, which is focused on describing the possible space of long-term, stable behaviors of a parameterized system, rather than the temporal trajectory of change given any particular parameterization. However, we believe that our findings are still useful in demonstrating that (i)
change is not inevitable in the presence of bias and (ii) not all biases engender the same evolutionary dynamics. This gives us confidence in the conclusions we can draw from (and the hypotheses we might generate with) such models.



    

      



\section{Conclusion}
In this paper, we have shown how different combinations of external forces can result in a similar evolutionary dynamics of the spread of change, at least in a setting with a strong categoricity bias. This implies that a similar evolutionary dynamics may actually underlie the actuation of changes from different sources. These results show us that although sound changes may have different sources, the resulting dynamics of propagation throughout a population may nonetheless be quite similar. We take this finding to be reassuring, given that empirical studies of many different changes show a similar trajectory.

At the same time, we have shown that not all external forces give rise to evolutionary dynamics where both stability and change are possible. Some intuitively plausible mechanisms, such as high social value associated with the speech of individuals, appear to be too noisy to have an effect when iterated over time in a speech community. This, too, is a positive result, because it demonstrates that the models in this general framework are not `doomed to success', but that there exist regions of the parameter space in which stability is possible. In this respect, population dynamics can provide at least a partial answer to the actuation problem and to the `non-phonologization problem' \citep{kiparsky2015phonologization} of why change does not automatically arise whenever its conditioning environment is present. Actuation is possible without production bias, and the dynamics of change driven by production bias are not unique.


\section*{Supplementary materials}

Simulation code to replicate and extend the simulations reported in this chapter can be found at this paper's OSF archive: \url{https://osf.io/b7sgq/}.

\section*{Acknowledgements}
Portions of this work were originally presented at the 2014 UC-Berkeley workshop `Sound Change in Interacting Human Systems'. We are grateful to that audience, as well as audiences at the Ohio State University and McGill University, editors Felicitas Kleber and Tamara Rathcke, an anonymous reviewer, and the participants in our 2013 and 2015 Linguistic Institute courses, for comments, suggestions, and inspiration. We alone remain solely responsible for any errors of fact or interpretation. This work was made possible in part by grants from the Fonds de recherche du Québec (\#183356) and the Canada Foundation for Innovation (\#32451) to M. Sonderegger. 

\DeclareRobustCommand{\disambiguate}[3]{#3}
\bibliographystyle{unified}
\bibliography{zotero}

\end{document}